# Fully Automatic Segmentation and Objective Assessment of Atrial Scars for Longstanding Persistent Atrial Fibrillation Patients Using Late Gadolinium-Enhanced MRI

Guang Yang [1,2†], Xiahai Zhuang [3†], Habib Khan [1], Shouvik Haldar [1], Eva Nyktari [1], Lei Li [4], Rick Wage [1], Xujiong Ye [5], Greg Slabaugh [6], Raad Mohiaddin [1,2], Tom Wong [1], Jennifer Keegan [1,2‡], David Firmin [1,2‡]

[1] Cardiovascular Biomedical Research Unit, Royal Brompton Hospital, SW3 6NP, London, U.K.

[2] National Heart and Lung Institute, Imperial College London, London, SW7 2AZ, U.K.

[3] School of Data Science, Fudan University, Shanghai, 201203, China.

[4] Dept. of Eng. Mechanics, School of NAOCE, Shanghai Jiao Tong University, Shanghai, 200240, China.

[5] School of Computer Science, University of Lincoln, LN6 7TS, Lincoln, U.K.

[6] Department of Computer Science, City University London, EC1V 0HB, London, U.K.

**Running Head:** Fully Automatic Scar Segmentation and Assessment for Longstanding AF Patients (78 Characters)

**Main Body Word Count:** 5000

**Abstract Word Count:** 207

**Figures:** 8

**Tables:** 2

[†] Corresponding authors: Guang Yang and Xiahai Zhuang

Emails: g.yang@imperial.ac.uk and zxh@fudan.edu.cn

[‡] Joint senior authors

Phone: (0044) 020 7352 8121; Fax: (0044) 020 7351 8699

Address: Cardiovascular Research Centre, Royal Brompton Hospital, SW3 6NP, London, U.K.

# ABSTRACT

**Purpose:** Atrial fibrillation (AF) is the most common cardiac arrhythmia and is correlated with increased morbidity and mortality. It is associated with atrial fibrosis, which may be assessed non-invasively using late gadolinium-enhanced (LGE) magnetic resonance imaging (MRI) where scar tissue is visualised as a region of signal enhancement. In this study, we proposed a novel fully automatic pipeline to achieve an accurate and objective atrial scarring segmentation and assessment of LGE MRI scans for the AF patients.

**Methods:** Our fully automatic pipeline uniquely combined: (1) a multi-atlas based whole heart segmentation (MA-WHS) to determine the cardiac anatomy from an MRI Roadmap acquisition which is then mapped to LGE MRI, and (2) a super-pixel and supervised learning based approach to delineate the distribution and extent of atrial scarring in LGE MRI.

**Results:** Both our MA-WHS and atrial scarring segmentation showed accurate delineations of cardiac anatomy (mean Dice = 89%) and atrial scarring (mean Dice =79%) respectively compared to the established ground truth from manual segmentation. Compared with previously studied methods with manual interventions, our innovative pipeline demonstrated comparable results, but was computed fully automatically.

**Conclusion:** The proposed segmentation methods allow LGE MRI to be used as an objective assessment tool for localisation, visualisation and quantification of atrial scarring.

**ABBREVIATIONS:** AF: Atrial Fibrillation; LA: Left Atrium; PV: Pulmonary Veins; SVM: Support Vector Machine; CV: Cross-Validation; ROI: Region of Interest.

# I. INTRODUCTION

## A. Background

Atrial fibrillation (AF) is the most common arrhythmia of clinical significance. It occurs when chaotic and disorganised electrical activity develops in the atria, causing muscle cells to contract irregularly and rapidly. It is associated with structural remodelling, including fibrotic changes in the left atrium [1] and can cause increased morbidity, especially stroke and heart failure. It also results in poor mental health, dementia, and increased mortality [2–4].

The electrical impulses that trigger AF frequently originate in the pulmonary veins (PV). Radio frequency ablation treatment aims to eliminate AF by electrically isolating the PV. However, the success rate for a single catheter ablation procedure is just 30-50% at 5 years follow-up [5, 6] and multiple ablations are frequently required.

The current clinical gold standard for assessment of atrial scarring is electro-anatomical mapping (EAM), performed during an electrophysiological (EP) study [7]. However, this is an invasive technique which uses ionising radiation the accuracy is suboptimal, with reported errors of up to 10 mm in the localisation of scar tissue [8, 9].

Late gadolinium enhancement (LGE) magnetic resonance imaging (MRI) is an established non-invasive technique for detecting myocardial scar tissue [10]. With this technique, healthy and scar tissues are differentiated by their altered wash-in and wash-out contrast agent kinetics, which result in scar tissue being seen as a region of enhanced or high signal intensity while healthy tissue is 'nulled'. While 2D breath-hold LGE MRI is well-established for ventricular imaging, there is a growing interest in imaging the thinner walled atria for identification of native and ablation scarring in AF patients [11–14]. This requires higher spatial resolution and contiguous coverage and data are best acquired as a 3D volume during free-breathing with diaphragmatic respiratory-gating. Atrial 3D LGE imaging has been used to: (1) assess patient suitability for AF ablation by identifying potential non-responders [12, 15–20], and (2) define the most appropriate ablation approach [16, 17, 21]. In addition, visualisation and quantification of native and post-ablation atrial scarring derived from LGE MRI has been used to guide initial and follow-up ablation procedures [17, 18, 22–25]. Histopathological studies in pigs have validated LGE MRI for the characterisation of AF ablation-induced wall injury [26].

Visualisation and quantification of atrial scarring requires objective, robust and accurate segmentation of the enhanced scar regions from the LGE MRI images. Essentially, there are two segmentations required: one showing the cardiac anatomy (geometry), particularly the LA wall and PV, the other delineating the enhanced scar regions. The former segmentation is required to rule out confounding enhanced tissues from other parts of the heart, e.g., the mitral valve and aorta, or the enhancement from non-heart structures while the latter is a prerequisite for visualisation and quantification. Segmentation of the atrial scarring from LGE MRI images is a very challenging problem. Firstly, the LA wall is very thin and scarring is hard to distinguish even by experienced expert cardiologists specialised in cardiac MRI. Secondly, residual respiratory motion, heart rate variability, low signal-to-noise ratio (SNR), and contrast agent wash-out during the long acquisition (current scanning time $\approx$10mins) frequently result in image quality being poor. Moreover, artifactual enhanced signal from surrounding tissues may result in a large number of false positives.

## B. Related Work

Oakes et al. [12] quantified the enhanced atrial scarring by analysing the intensity histogram of the manually segmented LA wall while Perry et al. [27] applied k-means clustering. A grand challenge for evaluation and benchmarking of various atrial scarring segmentation methods showed promising results [28]. Despite interest in these developed segmentation techniques, most of them have relied on manual segmentation of the LA wall and PV. This has several drawbacks: (1) it is a time-consuming task; (2) there are intra- and inter-observer variations; and (3) it is less reproducible for a multi-centre and multi-scanner study. Moreover, a number of studies have assumed a fixed thickness of the LA wall although there is no evidence that this is the case. Depending on the actual wall thickness, subsequent reorientation and interpolation of the MR images results in varying partial volume effects, which affect the apparent thickness of the LA wall. Inaccurate manual segmentation of the LA wall and PV can further complicate the delineation of the atrial scarring and its quantification can be error-prone. This could be one of the major reasons that there are currently on-going concerns regarding the correlation between atrial scarring identified by LGE MRI (enhanced regions) and 'gold standard' EAM (low voltage regions) [29].

The LA and PV would ideally be segmented from the cardiac and respiratory-gated LGE MRI dataset. However, this is difficult as normal tissue is 'nulled' and only scar tissue is seen with high signal. Other options are to segment them from a separately acquired breath-hold magnetic resonance angiogram (MRA) study [19, 30, 31] or from a respiratory and cardiac and respiratory gated 3D Roadmap acquisition, i.e., using a balanced steady state free precession (b-SSFP) sequence [32]. While MRA shows the LA and PV with high contrast, these acquisitions are generally un-gated and acquired in an inspiratory breath-hold. The anatomy extracted from MRA therefore can be highly deformed compared to that in the LGE MRI study. Although the 3D Roadmap acquisition takes longer to acquire, it is in the same respiratory phase as the LGE MRI and the extracted anatomy can be better matched. Cardiac anatomy has previously been defined by atlas based segmentation of MRA [31] and by using a statistical shape model [32] on 3D Roadmap data. Table 1 provides a summary of previously published methods on atrial scarring segmentation using LGE MRI.

## C. Our Contributions

In this paper, we present a novel fully automatic segmentation and objective assessment of atrial scarring for longstanding persistent AF patients scanned by LGE MRI. The LA chamber and PV are defined using a multi-atlas based whole heart segmentation (MA-WHS) method on Roadmap MRI images, which are acquired using a respiratory and cardiac gated 3D b-SSFP sequence. LA and PV geometry is resolved by mapping the segmented Roadmap anatomy to LGE MRI using the DICOM header data and is further refined by affine and nonrigid registration steps. The LGE MRI images are over-segmented by a novel Simple Linear Iterative Clustering (SLIC) based super-pixels method [33]. Then a support vector machines (SVM) based supervised classification is applied to segment the atrial scarring within the segmented LA and PV geometry. In this study, two validation steps have been performed—one for the LA chamber and PV segmentation and one for the atrial scarring segmentation—both against established ground truth from manual segmentations by experienced expert cardiologists specialised in cardiac MRI.

## II. MATERIALS AND METHODS

### A. Data Acquisition

Cardiac MR data were acquired on a Siemens Magnetom Avanto 1.5T scanner (Siemens Medical Systems, Erlangen, Germany).

Transverse navigator-gated 3D LGE MRI [11, 12, 34] was performed using an inversion prepared segmented gradient echo sequence (TE/TR 2.2ms/5.2ms) 15 minutes after gadolinium (Gd) administration when a transient steady-state of Gd wash-in and wash-out of normal myocardium had been reached [35]. LGE MRI images were scanned with a field-of-view 380×380mm$^2$ and reconstructed to 60–68 slices at 0.75×0.75×2mm$^3$.

Coronal navigator-gated 3D b-SSFP (TE/TR 1ms/2.3ms) Roadmap data were acquired with the following parameters: 80 slices at 1.6×1.6×3.2mm$^3$, reconstructed to 160 slices at 0.8×0.8×1.6mm$^3$, field-of-view 380×380mm$^2$, acceleration factor of 2 using GRAPPA, partial Fourier 6/8, acquisition window 125ms positioned within the subject-specific rest period, single R-wave gating, chemical shift fat suppression, flip angle 70°. Off resonant blood from the lungs arriving in the LA and PV can result in signal loss [36], which in our application, is minimised by using the shortest TE/TR possible. This was achieved by using non-selective RF excitation [37].

Both 3D LGE MRI and Roadmap data were acquired during free-breathing using a crossed-pairs navigator positioned over the dome of the right hemi-diaphragm with navigator acceptance window size of 5mm and CLAWS respiratory motion control [38]. The nominal acquisition duration was 204–232 cardiac cycles for 3D LGE MRI and 241 cardiac cycles for Roadmap assuming 100% respiratory efficiency.

## B. Patients

In agreement with the local regional ethics committee, cardiac MRI was performed in longstanding persistent AF patients between 2011–2013. The image quality of each dataset was scored by a senior cardiac MRI physicist on a Likert-type scale—0 (non-diagnostic), 1 (poor), 2 (fair), 3 (good) and 4 (very good)—depending on the level of SNR, appropriate TI, and the existence of navigator beam and ghost artefacts. Thirty seven cases with image quality greater or equal to 2 have been retrospectively entered into this study including 11 pre-ablation (included ~65% of pre-ablation cases) and 26 post-ablation scans (included ~92% of post-ablation cases).

## C. Multi-Atlas Whole Heart Segmentation (MA-WHS)

*MA-WHS Method*

A multi-atlas approach [39, 40] was developed to derive the whole heart segmentation of the Roadmap acquisition and then mapped to LGE MRI (Figure 1 (a)).

First we obtained 30 MRI Roadmap studies from the Left Atrium Segmentation Grand Challenge organised by King's College London [41] together with manual segmentations of the left atrium, pulmonary veins and appendages. In these, we further labelled the right and left ventricles, the right atrium, the aorta and the pulmonary artery, to generate 30 whole heart atlases. These 30 MRI Roadmap studies were employed only for building an independent multi-atlas dataset, which will then be used for segmenting our Roadmap studies that linked with the LGE MRI scans for the AF patients.

Let $I$ be the target image to be segmented, $\{(\boldsymbol{A}_a, \boldsymbol{L}_a)|a = 1, \ldots N\}$ be the set of atlases, where $N = 30$, $\boldsymbol{A}_a$ and $\boldsymbol{L}_a$ are respectively the intensity image and corresponding segmentation label image of the $a$-th atlas. For each atlas, MA-WHS performs an atlas-to-target registration, by maximizing the similarity between the images, to derive the set of warped atlases,

1) $$T_a = \arg\max_{T_a} \text{ImageSimilarity}(I, \boldsymbol{A}_a), \text{ and } \begin{cases} A_a = T_a(\boldsymbol{A}_a) \\ L_a = T_a(\boldsymbol{L}_a) \end{cases},$$

in which $T_a$ is the resulting transformation of the registration and $\{(A_a, L_a)|a = 1, \ldots N\}$ are respectively the warped atlas intensity image and corresponding segmentation result. Here, we employ the hierarchical registration for segmentation propagation, which was specifically designed for the whole heart MRI images and consists of three steps, namely the global affine registration for localisation of the whole heart, the local affine registration for the initialisation of the substructures, and the fully deformable registration for local detail refinement [42]. Image similarity metrics evaluate how similar the atlas and target image are. In this work we propose to use the spatially encoded mutual information (SEMI) method, which has been shown to be robust against intensity non-uniformity and different intensity contrast [43], that is

2) $$\text{ImageSimilarity}(I, A_a) = \{S_1, \dots, S_{n_s}\}$$

where $\{S_1, \dots, S_{n_s}\}$ are the SEMI and computed based on the spatially encoded joint histogram,

3) $$H_s(I, A_a) = \sum_{x \in \Omega} w_1(I(x)) w_2(A_a(x)) W_s(x).$$

Here, $w_1(I(x))$ and $w_2(A_a(x))$ are Parzen window estimation and $W_s(x)$ is a weighting function to encode the spatial information [43].

After the multi-atlas propagation, a label fusion algorithm is required to generate one final segmentation of the LA from the 30 propagated results,

4) $$L_I = \text{LabelFusion}(\{(A_1, L_1) \dots (A_N, L_N)\}).$$

The recent literatures have many new methods [44–51] on improving multi-atlas segmentation using sophisticatedly designed algorithms, which generally need to evaluate local similarity between patches from the atlases and the target image for local weighted label fusion,

5) $$L_I(x) = \text{argmax}_{l \in \{l_{bk}, l_{la}\}} \sum_a w_a(S(I, A_a, x)) \delta(L_a(x), l),$$

in which $l_{bk}$ and $l_{la}$ indicate the labels of the background and left atrium, respectively, and the local weight $w_a(\cdot) \propto S(\cdot)$ is determined by the local similarity $S(\cdot)$ between the target image and the atlas. $\delta(a, b)$ is the Kronecker delta function which returns 1 when $a = b$ and returns 0 otherwise.

For the LA segmentation, we propose to use the multi-scale patch based label fusion (MSP-LF). The multi-scale space theory can handle different level information within a small patch and has been applied to feature extraction/detection and image matching [40, 51–57]. The patches we compute from different scale spaces can represent the different levels of structural information, with low scale capturing local fine structure and high scale suppressing fine structure but providing global structural information of the image. To avoid increasing the computational complexity, we adopt the multi-resolution implementation and couple it with the MSP where the high-scale patch can be efficiently computed using a low-resolution image space. The local similarity between two images using the MSP measure is computed, as follows,

6) $$S_{\text{msp}}(I, A_a, x) = \sum_s S(I^{(s)}, A_a^{(s)}, x)$$

where $I^{(s)} = I * \text{Gaussian}(0, \sigma_s)$ is the target image from $s$ scale-space which is computed from the convolution of the target image with Gaussian kernel function with scale $s$. Here, we compute the local similarity in multi-scale image using the conditional probability of the images,

7) $$S(I^{(s)}, A_a^{(s)}, x) = p(i_x|j_x) = \frac{p(i_x, j_x)}{p(j_x)}$$

where $i_x = I^{(s)}(x)$ and $j_x = A_a^{(s)}(x)$ and the conditional image probability is obtained from the joint and marginal image probability which can be calculated using the Parzen window estimation [58].

For each patient, the Roadmap dataset was then registered to the LGE MRI dataset using the DICOM header data, and then refined by affine and nonrigid registration steps [43]. The resulting transformation was applied to the MA-WHS derived cardiac anatomy to define the endocardial LA boundary and PV on the LGE MRI dataset for each patient.

**D. Atrial Scarring Segmentation**

*Over-Segmentation by Simple Linear Iterative Clustering (SLIC) Based Super-Pixels*

We used a Simple Linear Iterative Clustering (SLIC) based super-pixel method [33] to over-segment LGE MRI images in order to separate potential enhanced atrial scarring regions from other tissues (Figure 1 (b)). Super-pixel algorithms group pixels into perceptually meaningful patches with similar size, which can be used to replace the regular pixel grid. In this study, we used a SLIC based super-pixel method, which has been successfully applied to solve various medical image analysis problems, e.g., [59, 60]. It has also demonstrated better segmentation accuracy and superior adherence to object boundaries, and it is faster and more memory efficient compared to other state-of-the-art super-pixels methods [33]. Based on local k-means clustering, the SLIC method iteratively groups pixels into super-pixels. The clustering proximity is estimated in both intensity and spatial domains that is

8) $$D = \sqrt{d_c^2 + \left(\frac{d_s}{S}\right)^2 m^2},$$

in which $d_c = \sqrt{(I_j - I_i)^2}$ measures the pixel intensity difference of a gray scale image and $d_s = \sqrt{(x_j - x_i)^2 + (y_j - y_i)^2}$ describes the spatial distance between each pixel and the geometric centre of the super-pixel. SLIC is initialised by sampling the target slice of the LGE MRI image into a regular grid space with grid interval of $S$ pixels. To speed up the iteration, SLIC limits the size of search region of similar pixels to $2S \times 2S$ around the super-pixel centre (namely local k-means clustering). In addition, parameter $m$ balances the weighting between intensity similarity $d_c$ and spatial proximity $d_s$.

*Support Vector Machines (SVM) Based Classification*

After SLIC segmentation, we proposed to use Support Vector Machines (SVM) to classify the over-segmented super-pixels into enhanced atrial scarring regions and non-enhanced tissues. SVM provide a powerful technique for supervised binary classification [61].

In order to train the SVM classifier, we built a training dataset containing enhanced and non-enhanced super-pixel patches (refer to Supporting Material Section A1). This has been done by (1) experienced expert cardiologists specialised in cardiac MRI performing manual mouse clicks to select the enhanced scar regions; (2) combining the mouse clicks and SLIC segmentation to label the enhanced super-pixels (refer to Supporting Material Figure S1); (3) applying morphological dilation to the segmented endocardial LA boundary and PV from MA-WHS to extract the LA wall and PV; (4) finding the overlapped regions of the LA wall and PV and the labelled enhanced super-pixels and (5) labelling the other super-pixels overlapped with LA wall and PV as non-enhancement (refer to Supporting Material Figure S2).

Instead of extracting texture or shape features of these labelled super-pixels, we computed the pixel-intensity based features to feed to the SVM classifier. Feature selection was done using minimum redundancy and maximum relevance method [62]. In this study, we applied the mutual information quotient scheme [62]. The selected features will be presented in the Results section and will be used for the further SVM based classification procedure. The parameters of the SVM with a RBF kernel were optimised using cross-validation with a grid search scheme [63] (refer to Supporting Material Section A4).

### E. Results Evaluation and Validation
*Evaluation and Validation of the MA-WHS*

One experienced cardiologist (>5 years' experience and specialised in cardiac MRI) manually segmented the endocardial LA boundary and labelled the PV slice-by-slice in the LGE MRI images for all the patients. A second senior cardiologist (>25 years' experience and specialised in cardiac MRI) confirmed the manual segmentation. The evaluation and validation of our MA-WHS has been done against this manual segmentation, which is assumed to be the ground truth. We used five metrics: DICE, JACCARD, PRECISION, HAUSDORFF distance [64] and Average Surface Distance (ASD) [65] (refer to Supporting Material Table S1).

*Ground Truth Definition of the Atrial Scarring*

We formed the ground truth of the enhanced atrial scarring on the LGE MRI images using the following steps:

(1) Steps (1)–(4) as listed in the 'SVM Based Classification' section.

(2) Once we extracted the enhanced super-pixels, they were combined to create a binary image for each slice, i.e., 1 for enhanced super-pixels and 0 for unenhanced.

(3) The binary image was overlaid on the original LGE MRI images and our cardiologists performed manual corrections to create the final boundaries (ground truth) of the enhanced atrial scarring. In so doing, we minimised the bias towards a better performance of the segmentation using classified super-pixels.

*Intra- and Inter-Observer Variances of the Manual Atrial Scarring Segmentation*

In this study, two cardiologists performed the manual mouse clicks based ground truth construction procedures in order to account for the inter-observer variance. Eight randomly selected persistent AF patients (4 pre- and 4 post-ablation cases) were entered for this test. In addition, one cardiologist performed the manual mouse clicks twice at two different time points (1 month in between) to estimate the intra-observer variance. The DICE metric was used to measure the intra- and inter-observer variances of the ground truth construction.

*Evaluation and Validation of the Fully-Automated Atrial Scarring Segmentation*

We evaluated our SVM based classification by: (i) leave-one-patient-out cross-validation (LOO CV), which provides an unbiased predictor and is capable of creating sufficient training data for studies with small sample size [66]; (ii) the cross validated classification accuracy, sensitivity, specificity, and average area under the receiver operating characteristic (ROC) curve (AUC) and (iii) the balanced error rate (BER) [67]. We also applied 10-fold CV to evaluate the robustness of our method when there are fewer manual labelled training datasets. Lastly, we divided our data into (a) a training/CV dataset (25 patients) and (b) an independent testing dataset (12 patients).

For the final atrial scarring segmentation we also performed results evaluation using DICE, JACCARD, PRECISION and NPV measurements. HAUSDORFF and ASD metrics were not applied because we have multiple discrete regions of the enhanced atrial scarring for each LGE

MRI volume. In addition, we also calculated the percentage of fibrosis extent [68, 19] (refer to Supporting Material Section A6).

*Comparison Study*

In order to demonstrate the efficacy of our method, we also compared it with two standard methods published in previous studies:

(a) Simple thresholding based method (Thr) [34]. The threshold value for each LGE MRI volume was chosen via empirical evaluation.

(b) Conventional standard deviation (SD) [12] based method (2, 4 and 6 SDs were tested).

These methods were selected as they have minimum parameter tuning and could be most accurately reproduced.

We compared the atrial scarring segmentation from each method against the ground truth using the LA and PV boundaries derived from our fully automatic MA-WHS segmentation. This was then repeated using the manually delineated LA and PV boundaries. The image intensities have also been normalised with respect to the intensities of the blood pool regions [28], which were extracted by a morphological erosion from the endocardial LA boundary.

## III. RESULTS

### A. Whole Heart Segmentation Results

Firstly, we have performed a study on various label fusion strategies. Figure 2 (a) shows the comparison results of the Dice scores using the different label fusion schemes, e.g., by majority vote (MV), by local weighted voting (LWV) [44], by joint label fusion (JLF) [50], by patch fusion one scale (PF), and by our proposed MSP. Our MSP was statistically significantly better than the other label fusion schemes ($p < 0.05$). Secondly, Figure 2 (b) shows the quantitative results of our MA-WHS method compared to the ground truth.

Figure 3 shows the reconstructed 3D surface mesh of the endocardial LA boundary and PV extracted using our MA-WHS method. We can observe that for most of the LA and the proximal PV, the segmentation accuracy is high; however, the segmentation in the more distal PV is more error-prone. In general, manual delineation always has more detailed segmentations of the PV structures and the extent of PV was under-estimated using our MA-WHS. However, for AF patients, it is only the very proximal regions that are of importance.

### B. Fully-Automated Atrial Scarring Segmentation Results

*Intra- and Inter-Observer Variances of the Ground Truth Construction*

Figure 4 demonstrates the intra- and inter-observer variances of the manual atrial scarring delineation. Results show a very good agreement (mean DICE scores ranging from 86%–92%) between the performances of the cardiologist (>5 years' experience) who did the manual mouse clicks twice (1 month between the two time-points). Compared to the performance of our second cardiologist (>25 years' experience), mean DICE scores range from 83%–91%. In summary, results of the DICE show that our atrial scarring ground truth construction (combining mouse clicks and super-pixels) has low intra- and inter-observer variance and that it is therefore valid for evaluation of the atrial scarring segmentation algorithms.

*Evaluation and Validation Results of the Fully-Automated Atrial Scarring Segmentation*

After minimum redundancy and maximum relevance based feature selection, 3 out of 16 features (minimum, mean and standard deviation) were selected and used in building the SVM model. Table 2 tabulates the SVM classification results of distinguishing enhanced atrial scarring regions from non-enhanced tissues. Using LOO CV, we obtained 88% accuracy and 0.16 BER. ROC analysis shows an AUC of 0.91 (refer to Supporting Material Figure S3). In terms of the final segmentation accuracy, we achieved mean DICE of 79%. We have also validated the proposed pipeline using 10-fold cross-validation that obtained similar accuracy. For the validation on the separated independent testing datasets, the fixed SVM model was blindly built. Compared to LOO CV on 37 datasets we achieved similar accuracy (86%) and sensitivity (92%), but lower specificity (64%) and mean DICE (71%) were obtained. This validation using separate testing datasets showed that our method can still perform well while fixing the classification model and using it to segment the new input data.

For the pre-ablation cases, using our fully automatic pipeline, the measured native fibrosis associated with AF was 26.9±11.2% (Figure 5 (a)). Compared to the pre-ablation cases, the fibrosis extent ratio was found to be 32.8±6.4% for the post-ablation cases (Figure 5 (a)). There was no significant difference found for the fibrosis extent derived using the ground truth segmentation of the atrial scarring (23.4±7.3% for the pre-ablation cases and 30.9±6.1% for the post-ablation cases). Comparing the fibrosis extent in our pre-ablation and post-ablation cases, both our fully automatic pipeline and the ground truth segmentation found significant differences in between (Figure 5 (a)). Bland-Altman analysis demonstrated that the fibrosis extent derived by our fully automatic segmentation pipeline has good agreement with the one derived using the ground truth segmentation (Figure 5 (b) and (c)).

Figure 6 shows the comparison results with simple thresholding and conventional standard deviation methods. Figure 6 (red bars) show that our method worked equally well in pre-ablation and post-ablation studies (median DICE score 80% for the post-ablation cases vs. median DICE score 76% for the pre-ablation cases and overall no significant difference by Wilcoxon rank-sum test, p=0.087). Overall, the atrial scarring segmentation results obtained using our method outperformed the simple thresholding and conventional standard deviation methods significantly (Figure 6).

Figure 7 shows our final atrial scarring segmentation results compared to the ground truth. The segmentation results have been derived from the LOO CV (i.e., training on 36 datasets and making prediction on the one dataset that has been left). For the first exemplar pre-ablation case (Figure 7 (a-c)), we can observe underestimated enhancement segmentation. Segmentation of the second pre-ablation case (Figure 7 (d-f)) shows clear accordance compared to the ground truth despite some overestimated region near the right inferior pulmonary vein. Segmentation of both post-ablation cases exhibit good agreement with the ground truth (Figure 7 (i) vs. (h) and Figure 7 (l) vs. (k)).

## IV. DISCUSSION

In this study, we developed a novel fully automatic segmentation pipeline to detect enhanced atrial scarring in LGE MRI images. Overall, results of this study offer compelling evidence that our fully automatic pipeline is capable of detecting enhanced atrial scarring from LGE MRI images acquired from a longstanding persistent AF cohort.

Segmentation of the atrial scarring from LGE MRI images is very challenging. This is not only because the atrial scarring is difficult to distinguish in the thin LA wall but also because the image quality can be poor due to motion artefacts, noise contamination and contrast agent wash-out during the long acquisition. Moreover, the enhancement from the surrounding tissues and enhanced blood flow are confounding issues for atrial scarring segmentation and result in increased false positives. However, most of these confounding enhancement regions can be distinguished subject to accurate heart anatomy delineation using our MA-WHS (Figure 7 (c), (f), (i) and (l)). Due to the subjective understanding of the LGE MRI images, our cardiologists may miss labelling some enhanced regions (double green arrows in Figure 7 (g)), but they can still be found using our supervised learning based fully automatic segmentation pipeline.

In our study, the size of the super-pixels was restricted by the LA wall thickness and $S$ was initialised to 4 pixels. We chose $m = 4$ based on visual inspections of the over-segmentation results. In order to evaluate the effect of $m$ we calculated the Dice scores between the ground truth LA and PV segmentations and the super-pixel derived LA and PV regions (Figure 8 (a)). Results showed that when $m = 4$ we achieved the highest mean Dice score and therefore the best adherence, yet no significant differences were found by setting $m = 1, 2, 8, 16,$ and 32. This can be attributed to the fact that we have a relatively small size of the super-pixels and the compactness term $m$ has less effect on the segmentation results.

Interestingly, the minimum redundancy and maximum relevance method has selected simple but effective features for our further SVM classification on SLIC segmented super-pixels. Three features were selected for the SVM classification, i.e., the mean, the standard deviation and the min of the super-pixels. The feature 'mean' corresponded to a simple thresholding on the super-pixel intensity values. The feature 'standard deviation' was selected because it quantified local intensity variations in the scar and healthy regions. The feature 'min' was selected as a strong discriminator due to the fact that in the enhanced atrial scarring regions the 'min' intensity values are much higher than the 'min' intensity values of the normal regions. However, the feature 'max' was not selected as a discriminator as for some labelled normal regions, there may be relatively high pixel intensities that could be false positives.

There are limitations of the current work. The fast and irregular heart rate in patients prior to ablation resulted in only 11 pre-ablation studies having good enough quality to be included in this study. Together with 26 post-ablation studies, our total number of datasets was limited to 37. To tackle the problem of having limited patient data, LOO CV was performed to achieve an unbiased predictor for limited datasets.

The comparison study reported here is also limited. A number of advanced techniques have been proposed such as unsupervised learning based clustering and graph-cuts based methods [28]. However, implementation of these is difficult as the fine-tuned hyper-parameters used are not always clearly described and the methodologies cannot be reproduced exactly. Moreover, our patient cohort is different from that in which these algorithms were optimised and tested. In this manuscript, we have therefore only compared our technique against the simple thresholding and conventional standard deviation based methods as these have fully-standard implementations. When compared to manual segmentation (ground truth) in post-ablation scans, these standard techniques gave median DICE of 38%–48% while our fully automatic technique achieved a median DICE of 80%. The results that we obtained here with the standard techniques are similar to those reported with these same techniques in the benchmarking study described in [28] while the latter score is similar to the best-performing methods reported in that same study. Moreover, we observed relatively large variance of the SD number above the mean signal intensity of the atrial blood pool regions (3.8±1.2 SDs derived using our fully automatic pipeline). This reflects not only on differences in gadolinium uptake in the scarred regions but also on correct setting of the inversion time. This might be one of the reasons that the method of using a fixed SD performed less well for our datasets. Of note is that in the benchmarking study the variances of all of the techniques tested are large while in our manuscript, the results are more consistent with a relatively small variance (boxplot in red as seen in Figure 6). This may be due to our patient cohort being more tightly defined while in the previous study, datasets were analysed from patients at multiple institutions using a variety of imaging protocols.

## V. CONCLUSIONS

To the best of our knowledge, this is the first study that developed a fully automatic segmentation pipeline for atrial scarring segmentation with quantitative validation for LGE MRI scans. The proposed pipeline has demonstrated an effective and efficient way to objectively segment and assess the atrial scarring. Our validation results have shown that both our MA-WHS and super-pixel classification based atrial scarring segmentation have obtained satisfactory accuracy. The current study was performed using real clinical data, and we can envisage an integration of our pipeline to clinical routines. In so doing, a patient-specific LA and PV geometry model and an objective atrial scarring segmentation can be obtained rapidly for individual AF patient without manual processing.

# ACKNOWLEDGEMENTS

This study was funded by the British Heart Foundation Project Grant (Project Number: PG/16/78/32402), the NIHR Cardiovascular Biomedical Research Unit, Royal Brompton Hospital & Harefield NHS Foundation Trust and Imperial College London, and the Chinese NSFC research fund (81301283) and the NSFC-RS fund (81511130090). Data were obtained during the NIHR Efficacy and Mechanism Evaluation Programme (Project Number: 12/127/127).

29

R.S., Marrouche, N.F.: New Magnetic Resonance Imaging-Based Method for Defining the Extent of Left Atrial Wall Injury After the Ablation of Atrial Fibrillation. J. Am. Coll. Cardiol. 52, 1263–1271 (2008).

# Table Captions

**Table 1:** Summary of the previously published methods for atrial scarring segmentation and ours.

**Table 2:** Quantitative evaluation of the atrial scarring segmentation. Three validation schemes were used (i.e., LOO CV, 10-fold CV, and training/CV with separate testing). The SVM based classification was evaluated using accuracy, sensitivity, specificity, BER and AUC. The final segmentation was evaluated against the ground truth using (Precision, NPV, Jaccard index and Dice score). Abbreviations: LOO–leave-one-patient-out; CV–cross-validation; BER–balanced error rate; AUC–area under curve; NPV–negative predictive value.

# Figure Captions

**Figure 1:** (a) Flowchart of the LA+PV segmentation via MA-WHS and its validation. (b) Flowchart of the fully automatic atrial scarring segmentation including atrial scarring ground truth construction, super-pixel and SVM classification based segmentation and leave-one-patient-out cross-validation. Abbreviations: LA+PV–left atrium and pulmonary veins; MAS–multi-atlas propagation based segmentation; MSP-LF–multi-scale patch based label fusion; WHS–whole heart segmentation.

**Figure 2:** (a) Comparison results (Dice scores of the WHS) of using different label fusion algorithms. ('*' = $p<0.05$ and '***' = $p<0.0005$; statistical significances were given by two-sample Wilcoxon rank-sum test). Abbreviations: MV–majority vote; LWV–local weighted voting; JLF–joint label fusion; PF–patch fusion one scale; MSP–multi-scale patch. (b) Quantitative evaluation (Dice score, Jaccard index, Precision, NPV, Hausdorff distance, and ASD) of the LA+PV segmentation using our MA-WHS method compared to the manual delineation of the cardiologists (i.e., ground truth). The mean and standard deviation (error bars) are shown. Abbreviations: LA+PV–left atrium and pulmonary veins; MA-WHS–multi-atlas whole heart segmentation; ASD–average surface distance.

**Figure 3:** Segmentation results (3D rendering) of two pre-ablation cases (a-c) and (d-f) and two post-ablation cases (g-i) and (j-l) are illustrated in four rows. (a), (d), (g) and (j) Manual delineated ground truth; (b), (e), (h) and (k) LA+PV segmentation via MA-WHS; (c), (f), (i) and (l) Hausdorff distance (in mm) calculated between the ground truth and the LA+PV segmentation. Abbreviations: LA+PV–left atrium and pulmonary veins; MA-WHS–multi-atlas whole heart segmentation.

**Figure 4:** Comparison of the manual delineations (ground truth) of the enhanced atrial scarring to demonstrate the inter- and intra-observer variances for 8 randomly selected patient cases. The mean and standard deviation (error bars) are shown. Abbreviations: GT_OP1_1: ground truth done by our first cardiologist at time point 1; GT_OP1_2: ground truth done by our first cardiologist at time point 2; GT_OP2_1: ground truth done by our second cardiologist.

**Figure 5:** (a) The percentage of fibrosis extent calculated using the ground truth segmentation (GT in blue boxplots) and using our fully automatic segmentation pipeline (SEG in purple boxplots) for both the pre-ablation cases and the post-ablation cases, respectively. ('*' = $p<0.05$ and 'n.s.' means no significant difference between two groups; statistical significances were given by two-sample Wilcoxon rank-sum test). Abbreviations: GT–ground truth segmentation; SEG–segmentation using our fully automatic pipeline. (b)-(c) Bland-Altman analysis of the measurements of fibrosis extent derived by using the ground truth segmentation ($FEP_{GT}$) and our fully automatic segmentation pipeline ($FEP_{SEG}$). Abbreviations: GT–ground truth segmentation; SEG–segmentation using our fully automatic pipeline; FEP–fibrosis extent percentage.

**Figure 6:** Comparison results with conventional atrial scarring segmentation methods using DICE. (a) Comparison results of the pre-ablation cases; (b) Comparison results of the post-ablation cases. ('*' = $p<0.05$, '**' = $p<0.005$, '***' = $p<0.0005$, and 'n.s.' means no significant difference between two groups; statistical significances were given by two-sample Wilcoxon rank-sum test). Abbreviations: Thr–simple thresholding based method with MA-WHS derived LA+PV; SD(x)–conventional standard deviation method (x=2, 4 and 6 SDs were tested) with MA-WHS derived LA+PV; Thr+M–simple thresholding based method with manual delineated LA+PV; SD(x)+M–conventional standard deviation method with manual delineated LA+PV; MA-WHS–multi-atlas whole heart segmentation; LA+PV–left atrium and pulmonary veins.

**Figure 7:** Final atrial scarring segmentation results of two pre-ablation cases (a-c) and (d-f) and two post-ablation cases (g-i) and (j-l). (a), (d), (g) and (j) Original LGE MRI images; (b), (e), (h) and (k) Ground truth of the atrial scarring segmentation; (c), (f), (i) and (l) Results of our fully automatic atrial scarring segmentation. Single red arrows in (a) and (d) show the enhancement of the AO wall. Double red arrows in (d) show the enhancement from non-heart tissue. Double red arrows (g) and (j) show the enhanced regions in other substructures of the heart or fat tissues surrounded. Single green arrow in (d) shows the enhanced artefacts of the mitral-valve. Double green arrows in (g) show some enhancement that might be missed in the ground truth labelling procedure, but found using our fully automatic segmentation. And double green arrows in (j) show the enhancement due to the navigator beam and blood flow. Abbreviations: LA–left atrium; AO–aorta; L–left; R–right.

**Figure 8:** (a) Illustration of calculating the Dice scores between the ground truth LA and PV segmentation (blue region) and the LA and PV segmentation derived from the SLIC super-pixels directly (green region); hexagons: represent dummy super-pixels, green hexagons: dummy super-pixels that have certain overlapping ratio with the ground truth segmentation and their boundaries defined the SLIC based segmentation (overlapping ratio was set to ≥20%); (b) Dice scores between the ground truth LA and PV segmentation and the LA and PV segmentation derived from the SLIC super-pixels directly by varying $m = 1, 2, 4, 8, 16, 32,$ and $64$. ('*' = $p<0.05$; statistical significances were given by two-sample Wilcoxon rank-sum test).

Table 1

| References | Subjects (number) | Cardiac Anatomy Segmentation (Modality) | Atrial Scarring Segmentation | Evaluation of Atrial Scarring Segmentation (Results: mean±std) |
|---|---|---|---|---|
| Oakes et al., 2009 | Human (81) | Manual Segmentation of LA Wall (LGE MRI) | 2-4 SD | Atrial Scarring Percentage (8±4, 21±6, 50±15) * |
| Knowles et al., 2010 | Human (7) | Semi-automatic Thresholding and Region Growing (MRA) | Maximum Intensity Projection | Atrial Scarring Percentage (31±10) † |
| Perry et al., 2012 | Human (34) | Manual Segmentation of LA Wall (LGE MRI) | k-means Clustering | Dice (81±11, Ground truth by manually selected thresholds) |
| Ravanelli et al., 2014 | Human (10) | Manual Segmentation of LA and PV in 3D (MRA) | 4 SD | Dice (60±21 Ground truth by a semi-automatic approach) ‡ |
| Karim et al., 2014 | Human (15) | Statistical Shape Model with Manual Correction (b-SSFP) | Graph-Cuts | Dice, ROC and Total Scar Volume ¶ |
| Tao et al., 2016 | Human (46) | Automatic Atlas Based Method with Level Set Refinement (MRA) | Maximum Intensity Projection | Qualitative Visualisation (N/A) |
| **Ours** | **Human (37)** | **Fully-automated Multi-Atlas Whole Heart Segmentation (b-SSFP)** | **Super-Pixel and SVM** | **Multiple Quantitative Metrics (Dice: 79±5)** |

\* Results (%) for mild (n=43), moderate (n=30) and extensive (n=8) enhancement cases.
† Moderate and extensive enhancement cases.
‡ The Dice score was calculated for an automated atrial scarring segmentation. The method was also evaluated using Bland-Altman analysis of the atrial scarring percentage (after skeletonisation) obtained from LGE MRI and EAM.
¶ Multiple Dice scores were calculated for various experimental settings, and they were reported by plotting the median Dice scores (around 80) with the minimum and the maximum.

Table 2

| Validation Method | | Accuracy (%) | Sensitivity (%) | Specificity (%) | BER | AUC | Precision (%) | NPV (%) | Jaccard Index (%) | Dice Score (%) |
|---|---|---|---|---|---|---|---|---|---|---|
| LOO CV (37 Patients) | | 88 | 90 | 79 | 0.16 | 0.91 | 81±9 | 99±1 | 65±6 | 79±5 |
| 10-Fold CV (37 Patients) | | 88 | 96 | 62 | 0.21 | 0.91 | 86±4 | 99±2 | 56±3 | 72±2 |
| Training/CV + Separate Testing | LOO CV (25 Patients) | 87 | 89 | 79 | 0.16 | 0.91 | 80±10 | 99±1 | 66±6 | 79±5 |
| | Separate Testing (12 Patients) | 86 | 92 | 64 | 0.22 | 0.88 | 77±7 | 99±1 | 56±7 | 71±7 |

Figure 1

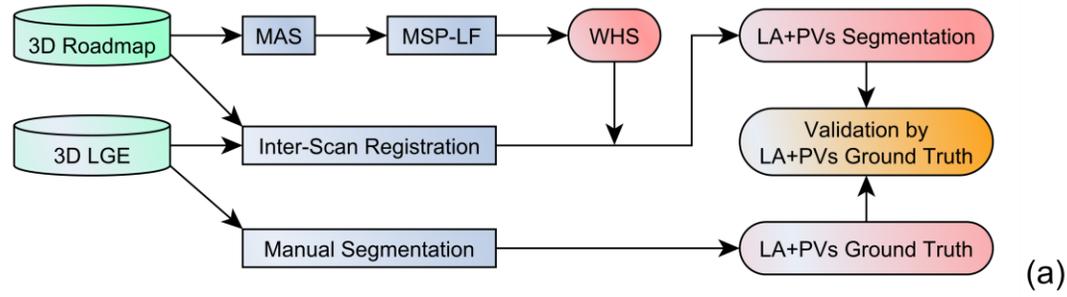

(a)

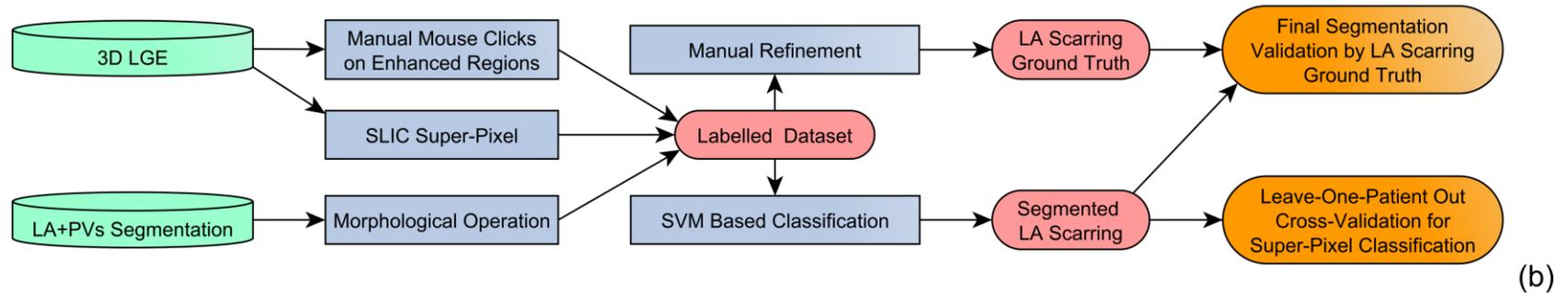

(b)

Figure 2

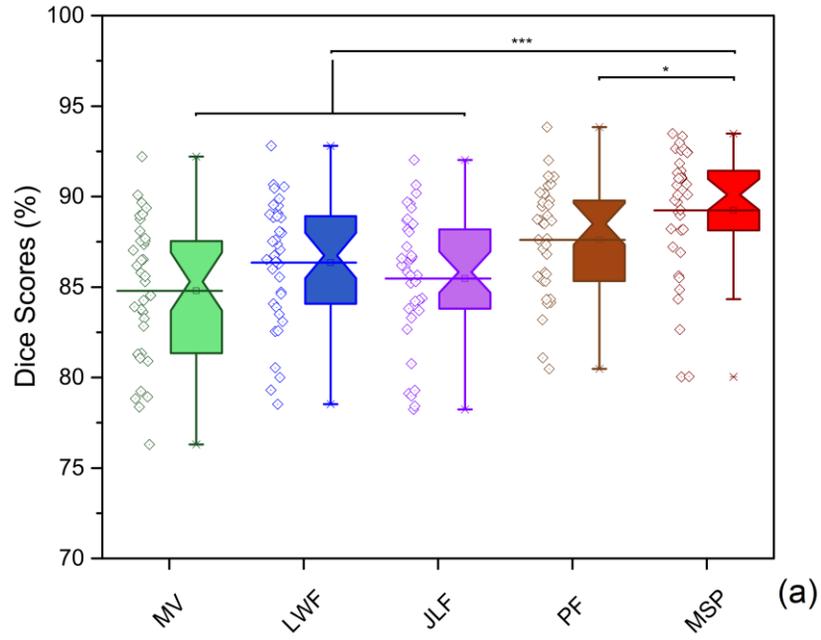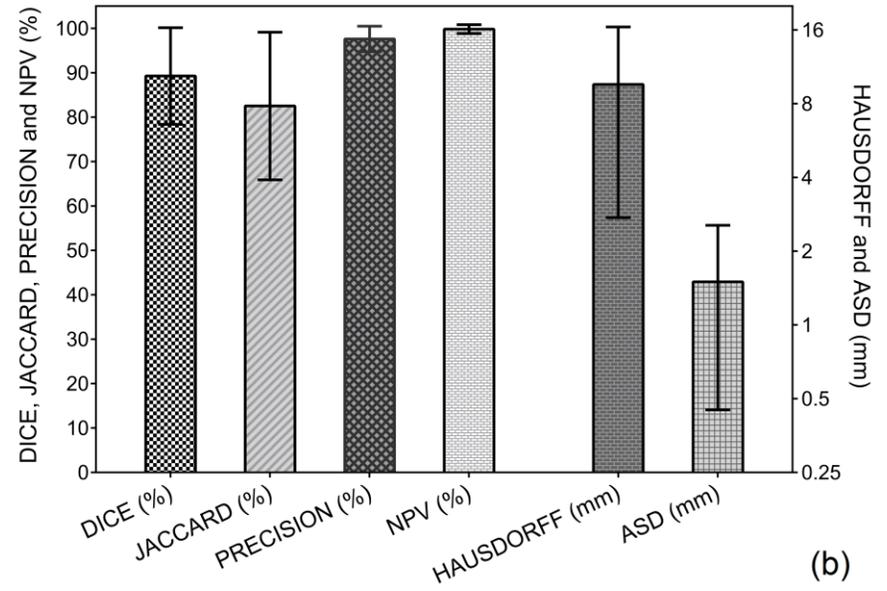

Figure 3

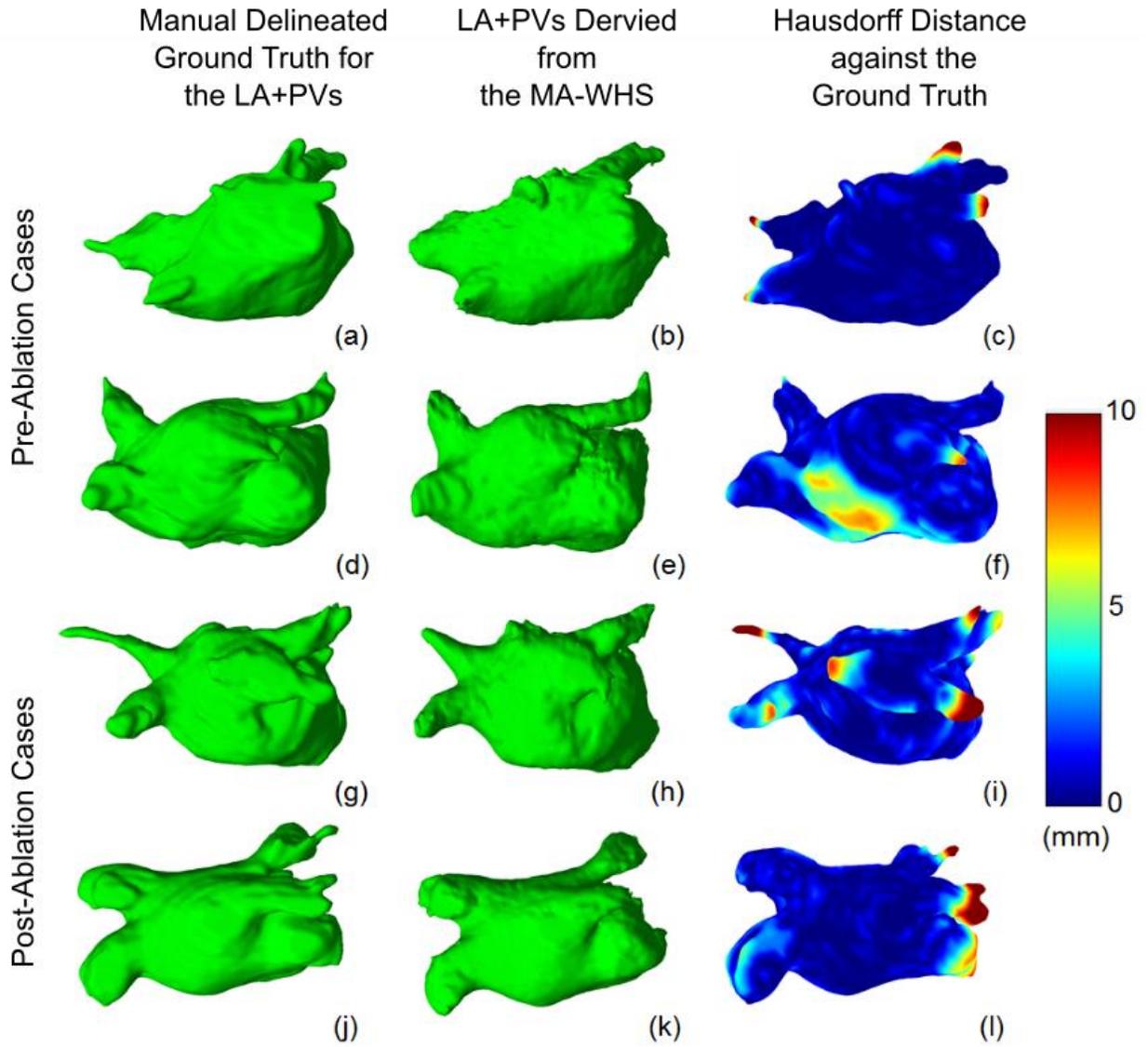

Figure 4

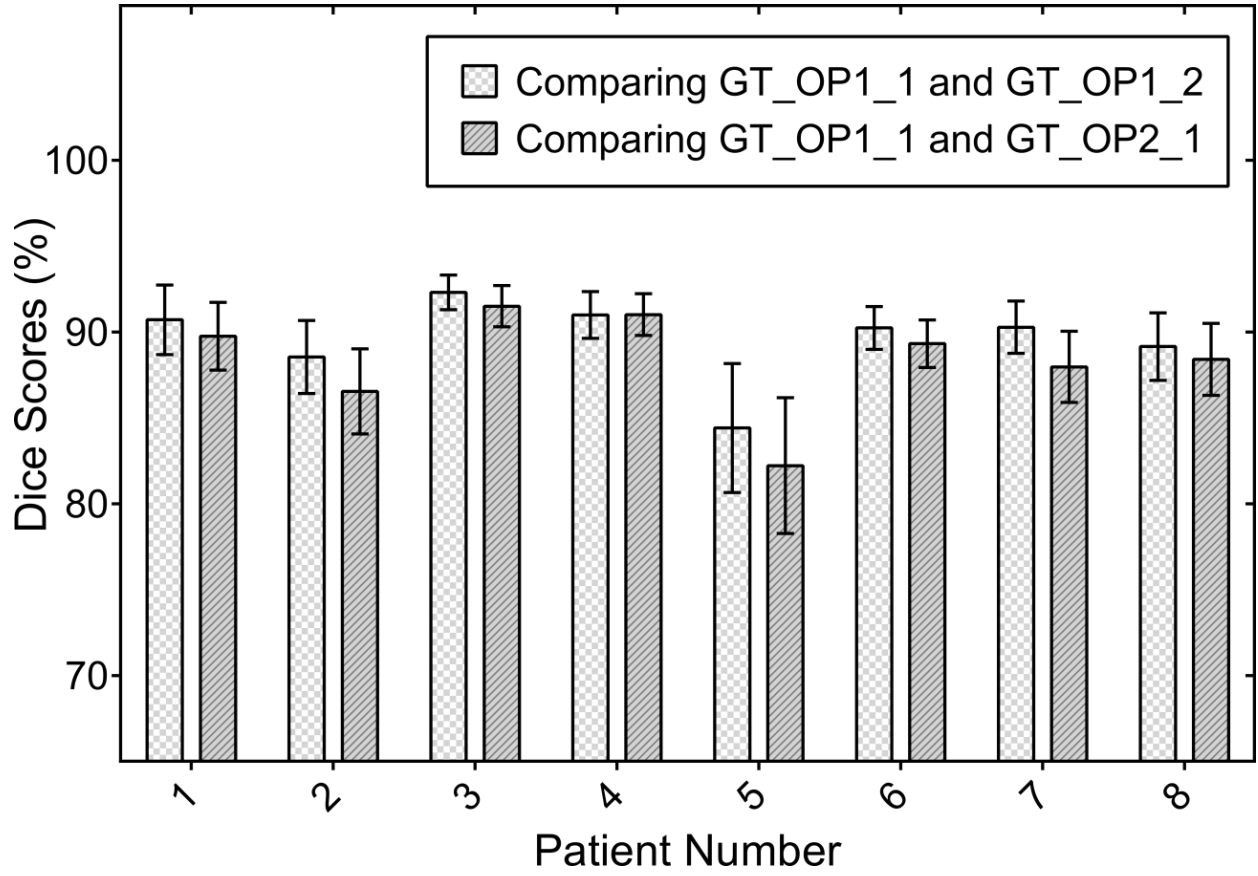

Figure 5

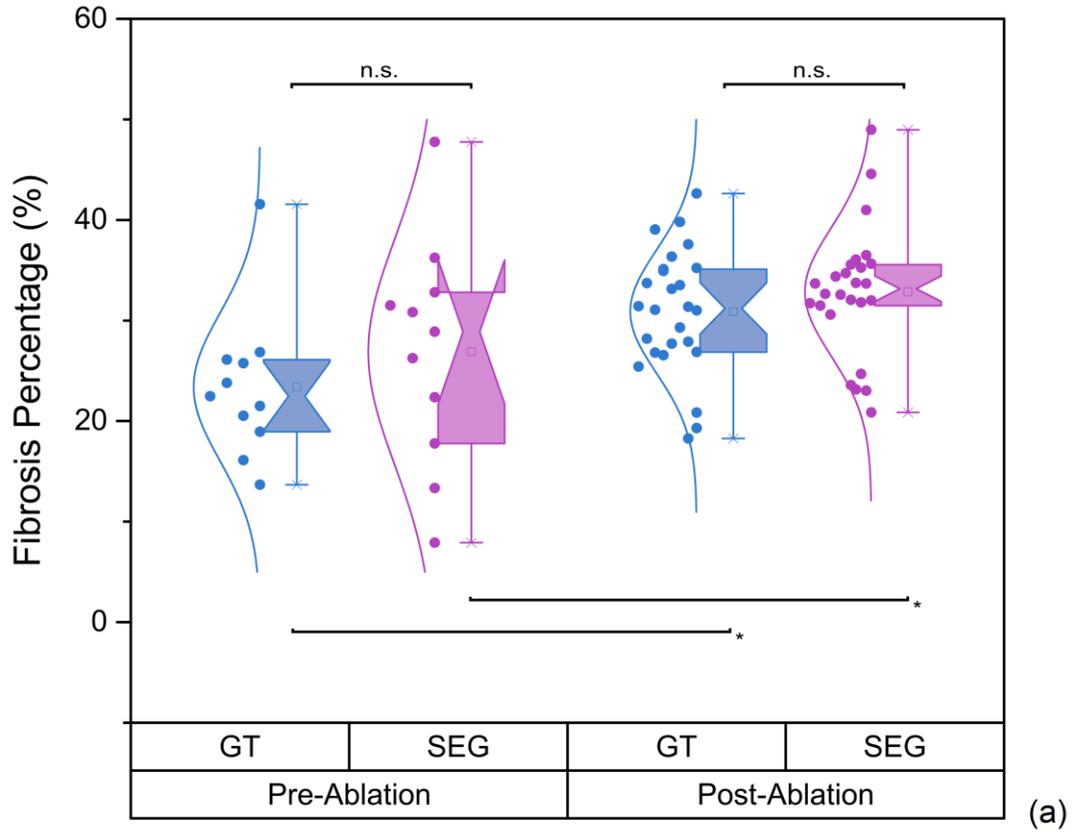

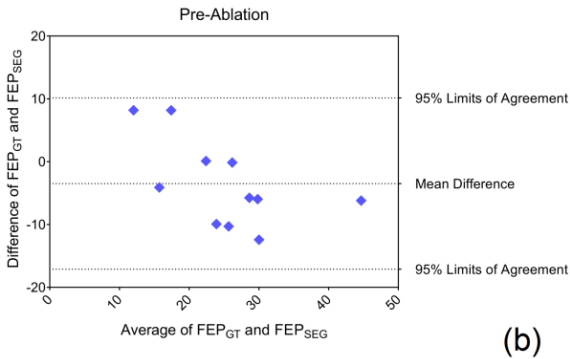
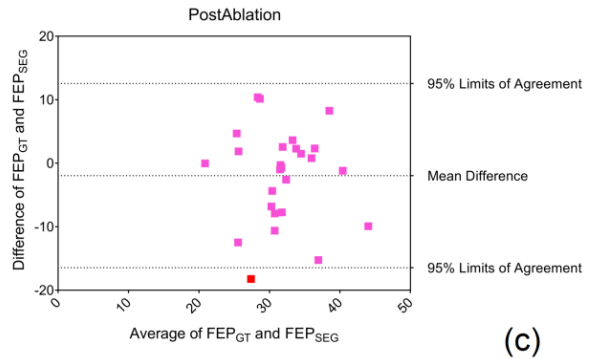

Figure 6

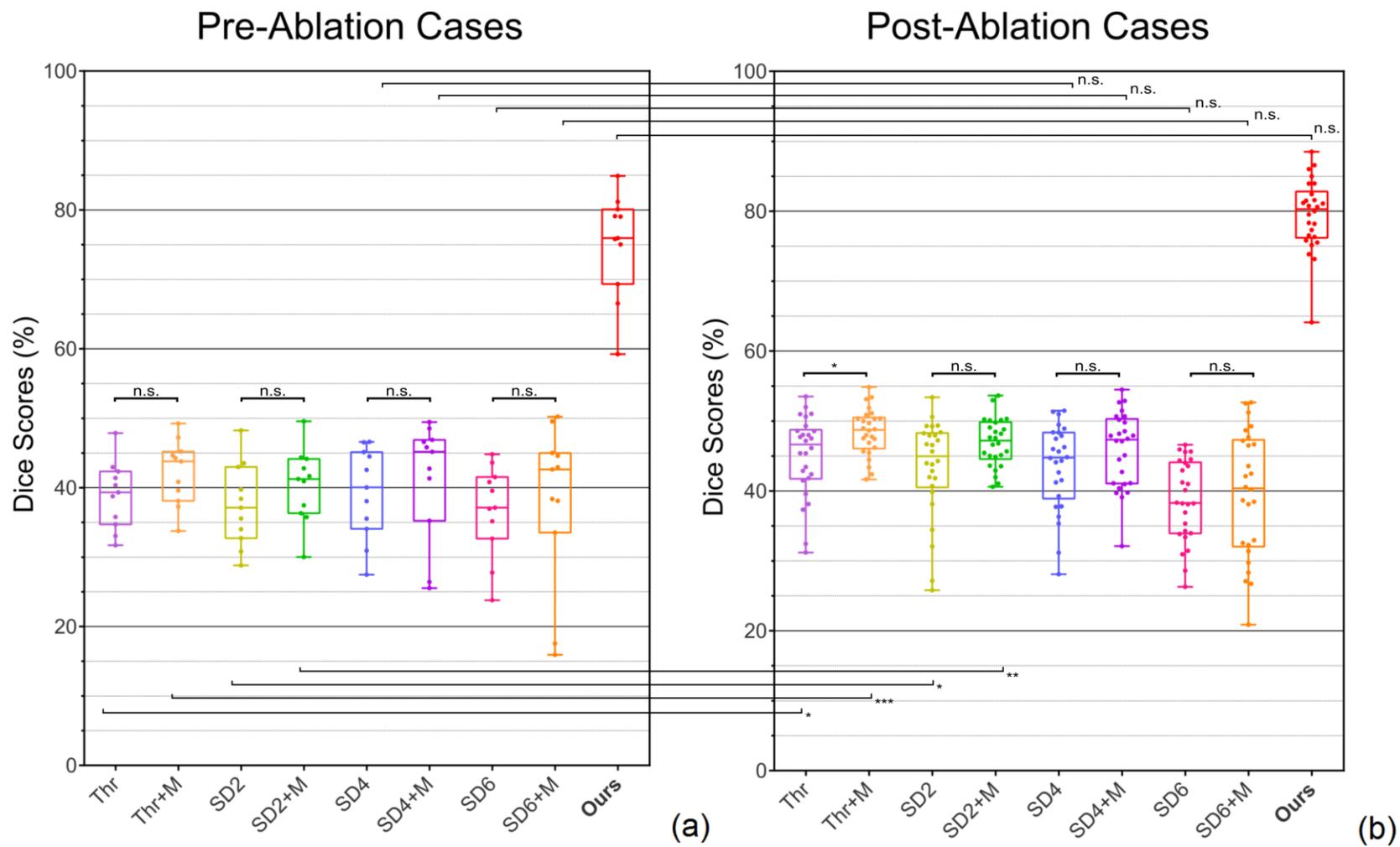

Figure 7

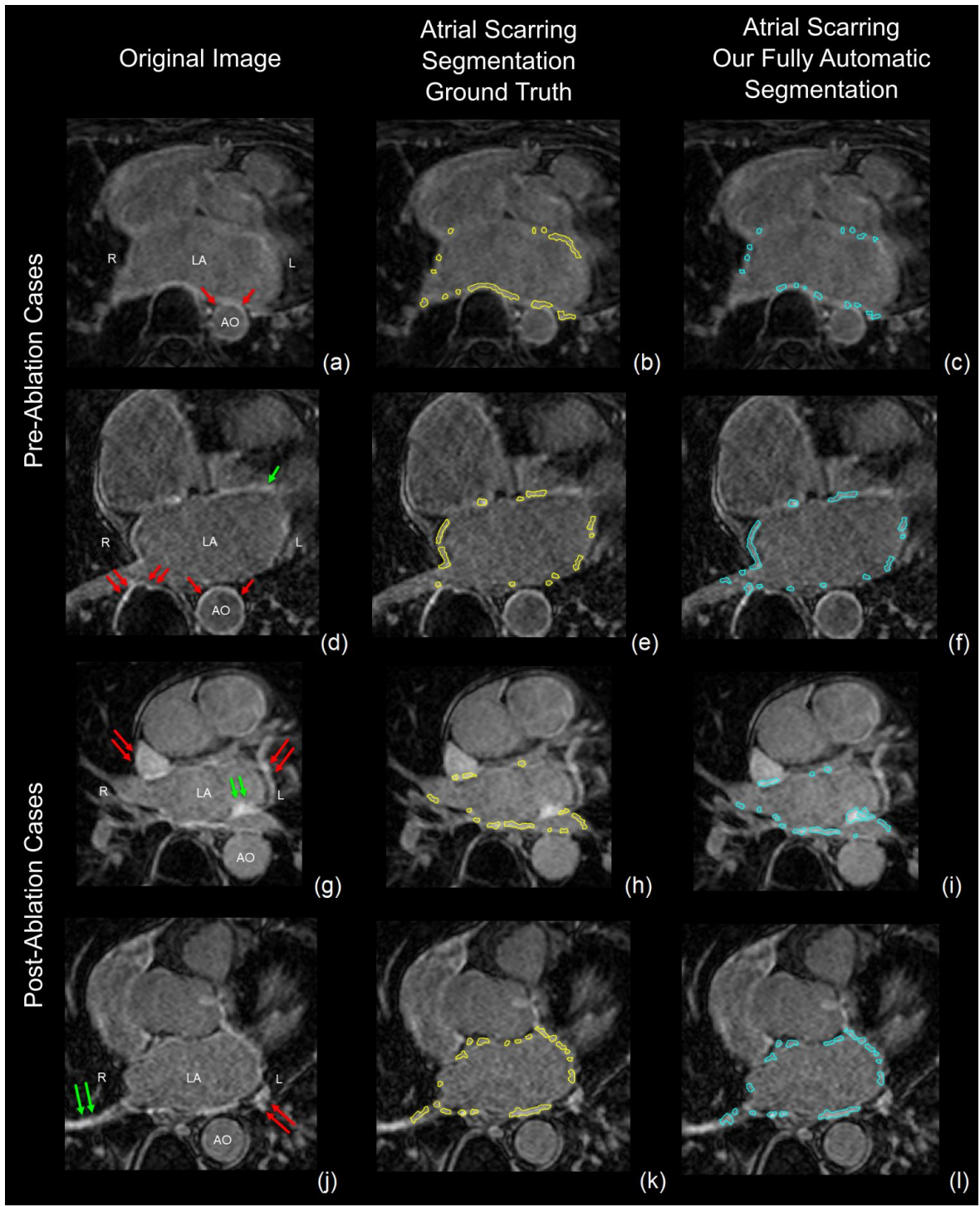

Figure 8

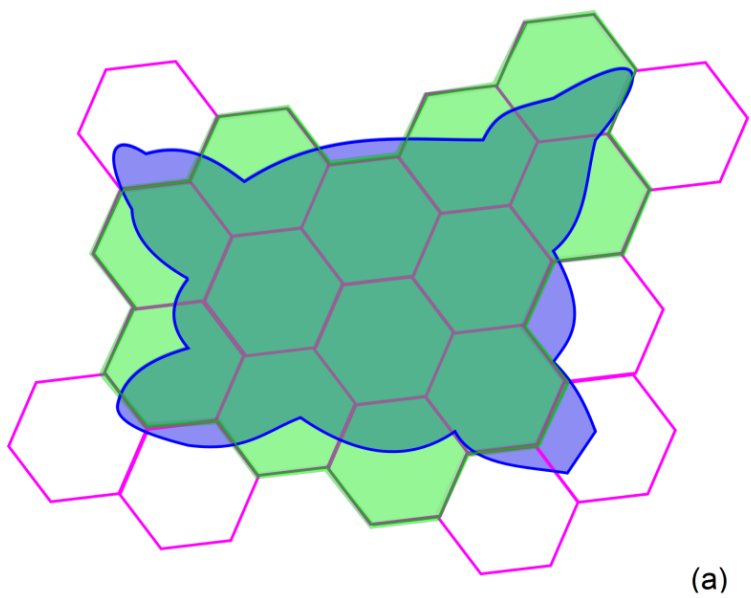 (a)

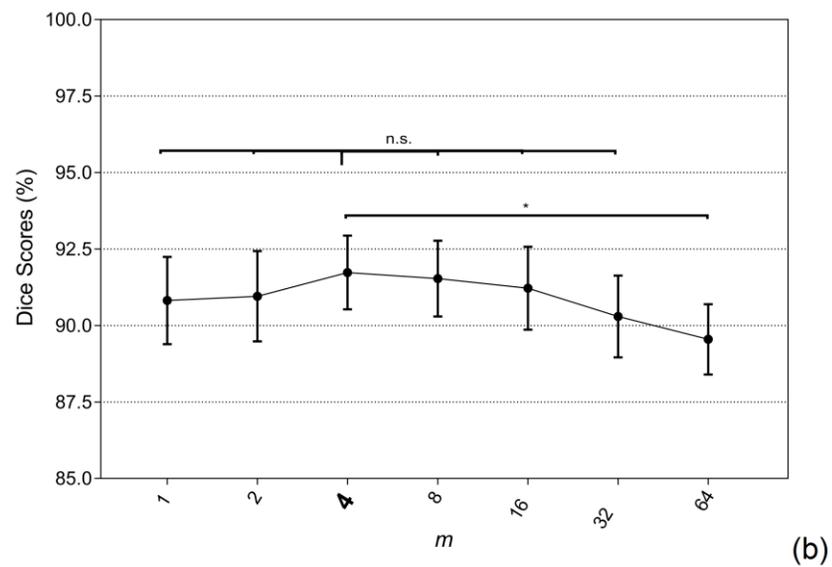 (b)

# Supporting Material

*A1. Training Dataset Construction*

Details of each step are given as following:

(1) Manual mouse clicks: Instead of manually drawing the boundaries of the enhanced atrial scarring regions, we asked experienced cardiologists specialised in cardiac MRI to perform manual mouse clicks on the LGE MRI images to label the regions that they believed to be enhanced (i.e., atrial scarring tissue). This is because manual boundary drawing of enhancement on the thin LA wall is a very challenging task and subject to large inter- and intra-observer variances. Mouse clicks on the enhancement regions are much easier and much more efficient. The manual mouse clicks were done on the original LGE MRI images without the super-pixel grid overlaid. This is because: (a) the mouse clicks will not be biased by super-pixel patches and (b) the super-pixel grid may reduce the visibility of the enhancement on LGE MRI images.

(2) The coordinates of the mouse clicks were used to select the enhanced super-pixels. Because our cardiologists performed the mouse clicks on the original LGE MRI images without having prior knowledge about the super-pixels, we asked them to have relatively dense mouse clicks. These mouse clicks will ensure all the enhanced regions can be included, but only one mouse click will be taken into account if multiple clicks dwell in the same super-pixel.

(3) The endocardial LA boundary and PV were extracted using our MA-WHS method. We then applied a morphological dilation to extract the LA wall and PV assuming the LA wall thickness is approximately 3mm [1, 2], and also take into account that the super-pixel size is still large enough to extract statistics of the grouped pixel intensities. The blood pool regions were extracted by a morphological erosion (5mm) from the endocardial LA boundary. And the pixel intensities were normalised according to the mean and standard deviation of the blood pool intensities [3].

(4) We masked the selected enhanced super-pixels [derived from step (2)] using the LA wall and PV segmentation. Only the super-pixels having a defined overlap with the LA wall and PV segmentation were selected as enhancement for building the training data (overlapping ratio was set to $\geq 20\%$). Other super-pixels (overlapping ratio $<20\%$) were discarded as they were considered as enhancement from other substructures of the heart (such as the mitral valve and aorta) but not enhancement of the LA wall and PV. Although we assumed that the LA wall thickness is 3mm, our enhanced super-pixels are not restricted to this wall thickness.

(5) The other super-pixels overlapped with the LA wall and PV but not selected as enhancement were considered as non-enhancement (overlapping ratio was set to ≥20%).

By performing the five steps described above, we constructed a training dataset that contains super-pixels labelled either enhancement or non-enhancement within the LA wall and PV.

## A2. SLIC Super-Pixel Over-Segmentation And Mouse Clicks

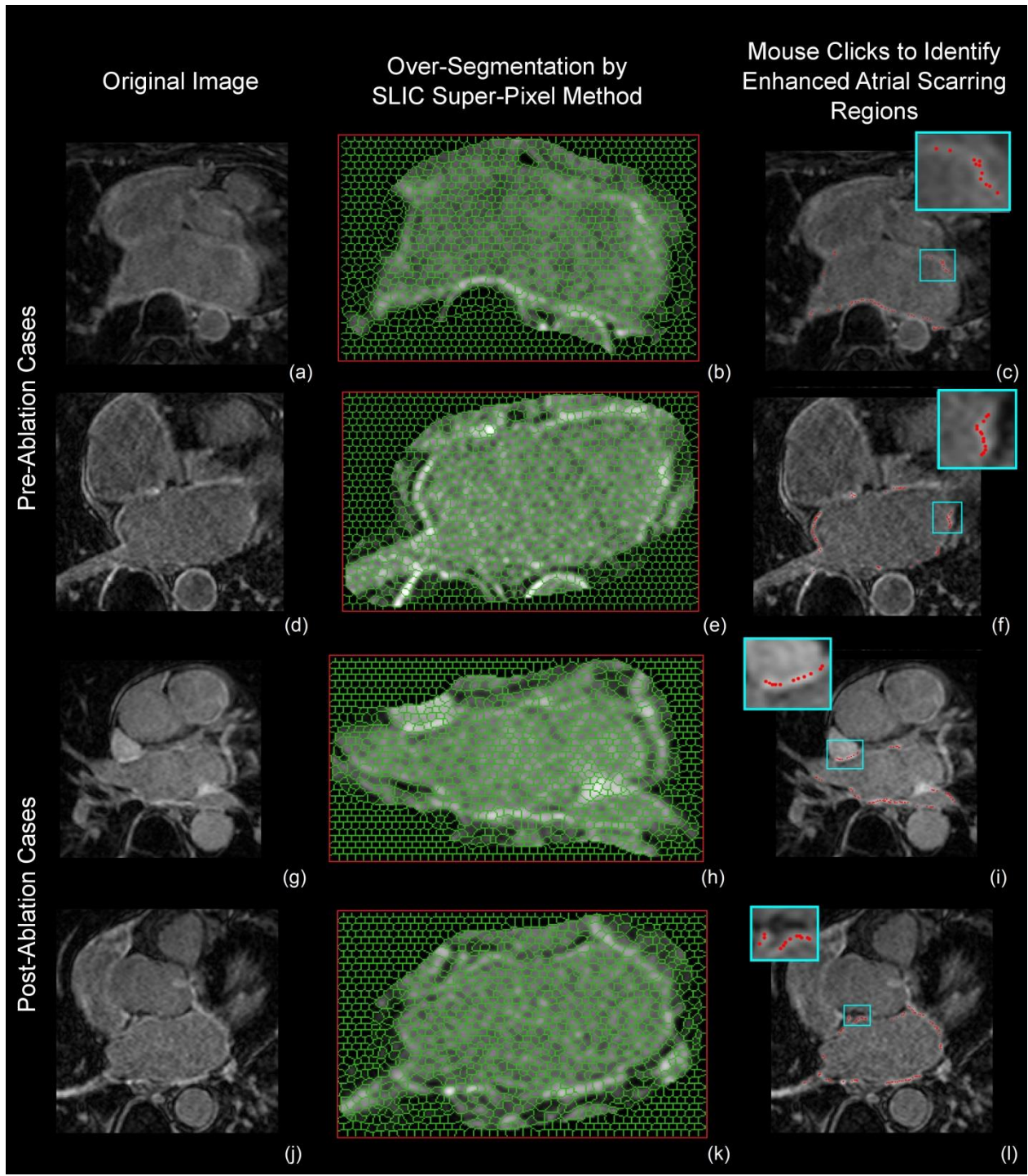

**Supporting Material Figure S1:** Super-pixel and mouse clicks based ground truth and trainings data construction. Two pre-ablation cases (a-c) and (d-f) and two post-ablation cases (g-i) and (j-l) are illustrated in four rows. (a), (d), (g) and (j) Original LGE MRI images; (b), (e), (h) and (k) Super-pixel results of the ROIs containing LA+PV (zoomed-in to show the details of the super-pixels boundaries); (c), (f), (i) and (l) Manual mouse clicks (red points) from one of our cardiologists (zoomed-in details of the mouse clicks are shown in cyan boxes). Abbreviations: LA+PV–left atrium and pulmonary veins; ROIs –region of interest.

## A3. MA-WHS Results and Labelled Training Dataset

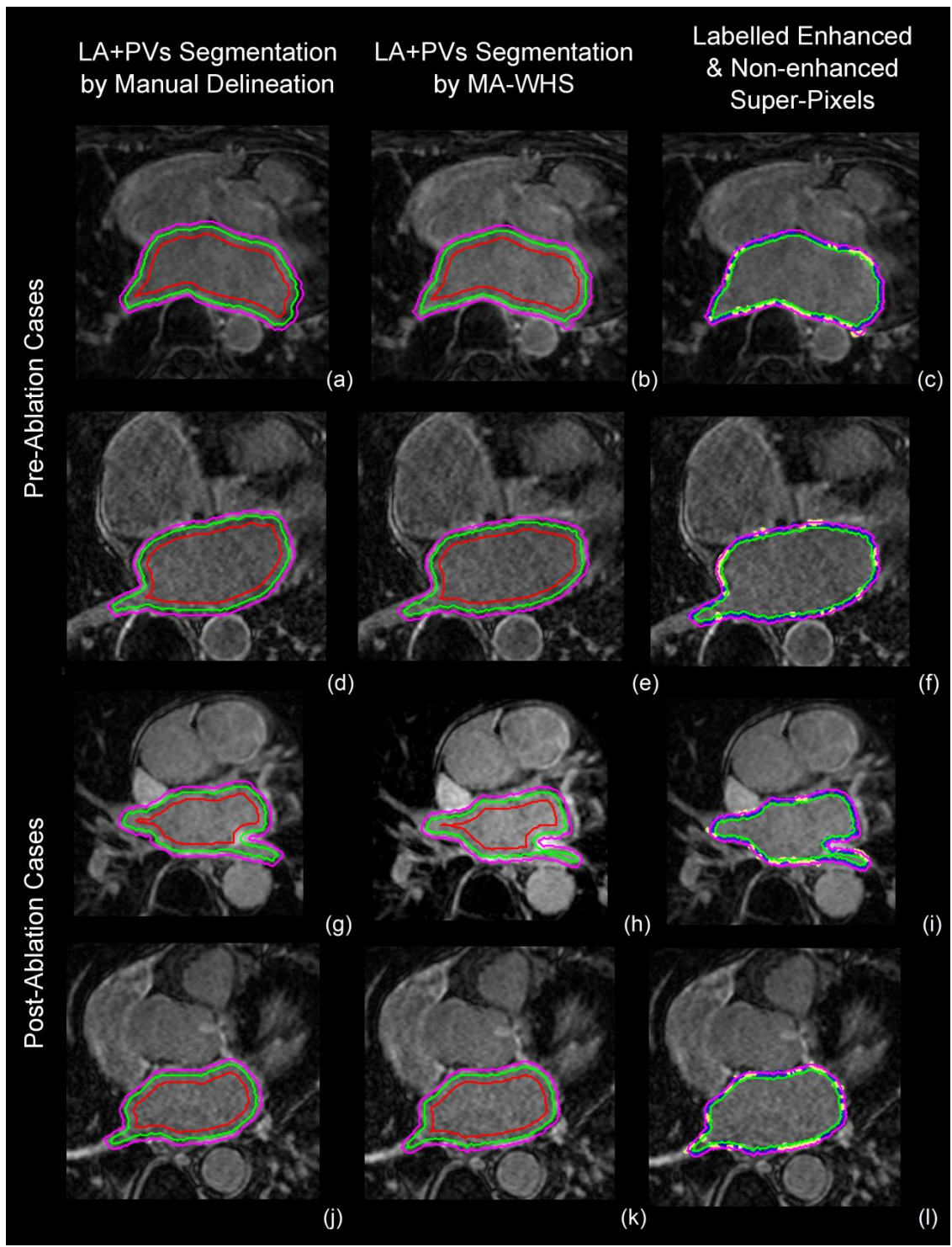

**Supporting Material Figure S2:** Segmented LA+PV and the constructed training dataset. Two pre-ablation cases (a-c) and (d-f) and two post-ablation cases (g-i) and (j-l) are illustrated in four rows (the same patients as seen in Supporting Material Figure S1). (a), (d), (g) and (j) Extracted LA wall and PV (regions between green and magenta curves) and blood pool (regions within the red curve) after morphological operations using manual delineation from one of our cardiologists; (b), (e), (h) and (k) Extracted LA wall and PV (regions between green and magenta curves) and blood pool (regions within the red curve) after morphological operations using MA-WHS; (c), (f), (i) and (l) Labelled training dataset for further classification (super-pixels labelled as the enhanced atrial scarring in yellow, and non-enhanced super-pixels in blue). Abbreviations: MA-WHS–multi-atlas whole heart segmentation; LA+PV–left atrium and pulmonary veins.

*A4. Support Vector Machines (SVM)*

After SLIC segmentation, we proposed to use Support Vector Machines (SVM) to classify the over-segmented super-pixels into enhanced atrial scarring regions and non-enhanced tissues. SVM provide a powerful technique for supervised binary classification [4]. SVM predictions depend on a subset of training data (i.e., the support vectors), and find the hyperplane with largest margin between the two classes [5]. This is obtained by solving the following optimisation problem,

1) $$\min_{w,b,\zeta} \frac{1}{2} w^T w + \rho \sum_{i=1}^{l} \zeta_i ,$$

$$\text{subject to } y_i(w^T \emptyset(x_i) + b) \geq 1 - \zeta_i \text{ and } \zeta_i \geq 0,$$

in which $(x_i, y_i), i = 1,2, \dots, l$ is the instance-label pairs of the given training dataset [6]. Here $\langle w, x \rangle + b = 0$ defines the separating hyperplane for $b \in \mathbb{R}$ is real. Furthermore, $L_1$-norm based formulation of the soft margins was applied by adding slack variables $\zeta_i$ and a penalty parameter $\rho$, which is known as the box constraint for the soft margin. In addition, $K(x_i, x_j) \equiv \emptyset(x_i)^T \emptyset(x_j)$, is called the kernel function. In this study, we used a nonlinear Gaussian Radial Basis Function (RBF) kernel $K(x_i, x_j) = \exp(-\gamma \|x_i - x_j\|^2)$ with scaling-factor, $\gamma > 0$, to map feature vectors into a nonlinear feature space where an optimal hyperplane was constructed to separate two different classes, i.e., enhancement and non-enhancement. The parameters of the SVM with a RBF kernel (i.e., $\rho$ and $\gamma$) were optimised using cross-validation with a grid search scheme [6]. In this study we firstly used coarse $11 \times 11$ 'grid' using $\rho = 2^{-10}, 2^{-8}, \dots, 2^8, 2^{10}$ and $\gamma = 2^{-10}, 2^{-8}, \dots, 2^8, 2^{10}$, and then with a fine $3 \times 3$ 'grid' $\rho = 2^8, 2^{8.5}, 2^9$ and $\gamma = 2^2, 2^{2.5}, 2^3$. The optimisation showed that the best classification was achieved when $\rho = 2^{8.5}$ and $\gamma = 2^{2.5}$.

*A5. Summary of the Quantitative Evaluation Methods*

**Supporting Table S1:** Summary of the quantitative evaluation methods. $F_{Manual}$: ground truth segmentation; $F_{Auto}$: automatic segmentation; $|\bullet|$: the number of pixels assigned to the segmentation; T: the total number of pixels; $P_{Manual} = \{p_{m1}, \cdots, p_{mn}\}$ and $P_{Auto} = \{p_{a1}, \cdots, p_{an}\}$: two finite point sets of the two segmented contours (using the ground truth segmentation and automatic segmentation); $\|\bullet\|$: $L_2$ norm; sup: supremum and inf: infimum. For the MA-WHS method we used all the six evaluation metrics and for the final atrial scarring segmentation we employed DICE, JACCARD, PRECISION, and NPV.

| Evaluation Metrics | Definition | MA-WHS | Atrial Scarring Segmentation |
|---|---|---|---|
| Dice score | $DICE = \frac{2 \times |F_{Manual} \cap F_{Auto}|}{|F_{Manual}| + |F_{Auto}|}$ | • | • |
| Jaccard index | $JACCARD = \frac{|F_{Manual} \cap F_{Auto}|}{|F_{Manual} \cup F_{Auto}|}$ | • | • |
| Precision | $PRECISION = \frac{|F_{Manual} \cap F_{Auto}|}{|F_{Auto}|}$ | • | • |
| Negative Predictive Value | $NPV = \frac{T - |F_{Manual} \cup F_{Auto}|}{T - |F_{Auto}|}$ | • | • |
| Hausdorff distance | $HAUSDORFF(P_{Manual}, P_{Auto}) = \max(d(P_{Manual}, P_{Auto}), d(P_{Auto}, P_{Manual}))$ where $d(P_{Manual}, P_{Auto}) = \sup_{p_m \in P_{Manual}} \inf_{p_a \in P_{Auto}} \|p_m - p_a\|$ $d(P_{Auto}, P_{Manual}) = \sup_{p_a \in P_{Auto}} \inf_{p_m \in P_{Manual}} \|p_m - p_a\|$ | • | |
| Average Surface Distance | $ASD = \frac{1}{2} \left( \frac{\sum_{p_m \in P_{Manual}} \min_{p_a \in P_{Auto}} \|p_m - p_a\|}{\sum_{p_m \in P_{Manual}} 1} + \frac{\sum_{p_a \in P_{Auto}} \min_{p_m \in P_{Manual}} \|p_m - p_a\|}{\sum_{p_a \in P_{Auto}} 1} \right)$ | • | |

We used five metrics: DICE, JACCARD, PRECISION, HAUSDORFF distance [7] and Average Surface Distance (ASD) [8]. DICE, JACCARD, PRECISION, and Negative Predictive Value (NPV) measure the overlap (in %) between two segmentations. JACCARD is numerically more sensitive to mismatch when there is reasonably strong overlap than DICE or PRECISION. The higher the values of DICE, JACCARD, NPV, and PRECISION, the better the overall performance of the segmentation will be. HAUSDORFF and ASD measure the boundary distance (in mm) between two contours of segmentation. The lower the values of HAUSDORFF and ASD the better agreement between manual delineation and fully automatic segmentation.

*A6. Fibrosis Extent Measurement*

In addition, the fibrosis extent measurement of the atrial scarring is an important imaging biomarker for predicting the outcome of the AF treatment. Previous studies used atrial scarring volume to measure the fibrosis extent [9–11]. In this study, instead of reporting the absolute atrial scarring volume, we used the percentage of fibrosis extent [12, 13] that is calculated as the ratio between the segmented atrial scarring volume ($V_{SEG}$) and the total LA wall volume ($V_{WALL}$)

$$\text{FEP}_{SEG} = 100 \times \frac{V_{SEG}}{V_{WALL}}, \quad (2)$$

and we also compare this with the percentage of fibrosis extent that is derived using the manual ground truth segmentation of the atrial scarring

$$\text{FEP}_{GT} = 100 \times \frac{V_{GT}}{V_{WALL}}. \quad (3)$$

*A7. ROC Analysis*

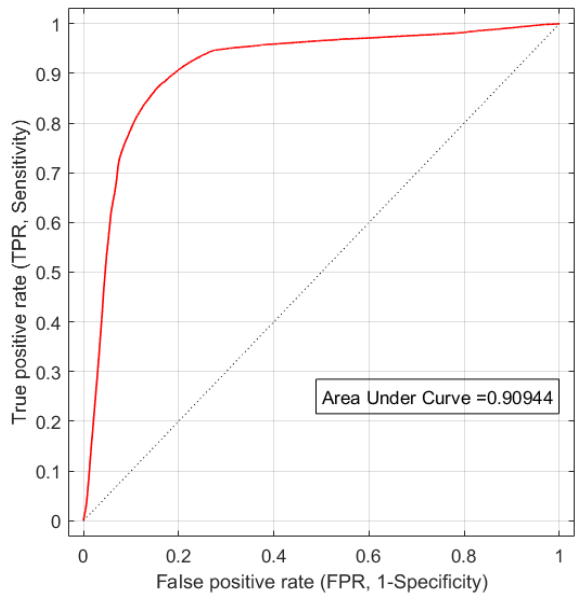
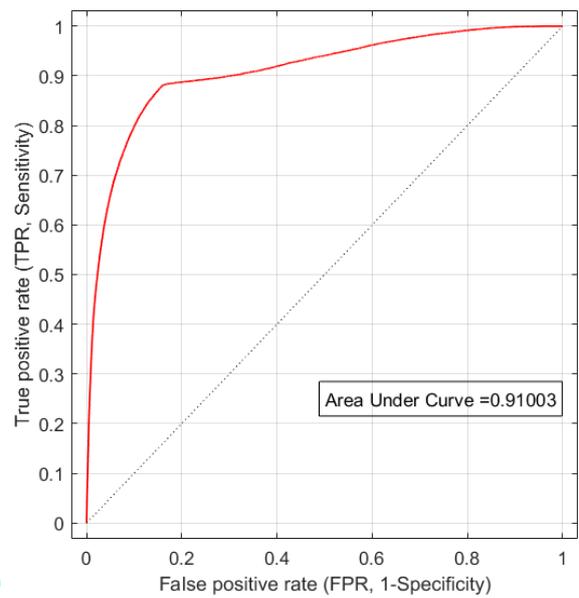
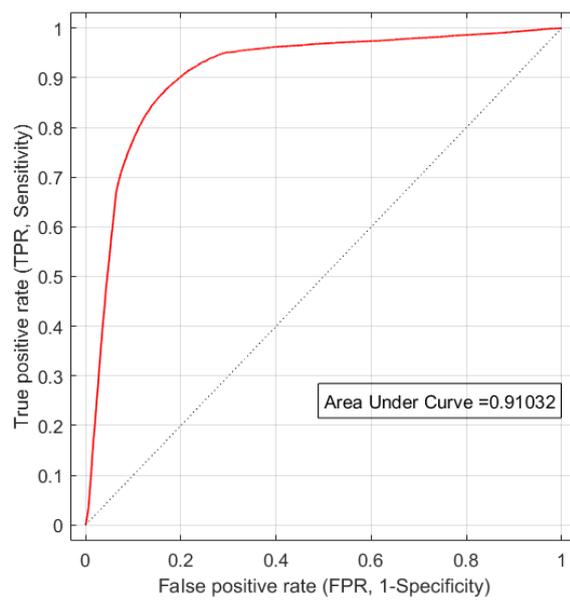
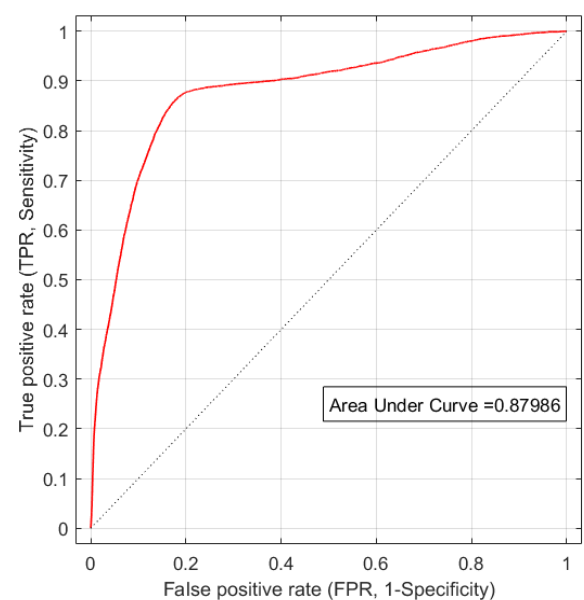

**Supporting Material Figure S3:** ROC curves of the different cross-validation methods. (a) Leave-one-patient-out cross validation (LOO CV) for 37 patients; (b) 10-fold cross validation for 37 patients; (c) LOO CV for 25 patients of the SVM model training; (d) Separate testing for 12 patients. Abbreviations: ROC–receiver operating characteristic; LOO CV–leave-one-patient-out cross-validation; SVM–support vector machine.

# REFERENCES FOR THE SUPPORTING MATERIAL